\documentclass[letterpaper, 10 pt, conference]{ieeeconf}  

\IEEEoverridecommandlockouts                              

\makeatletter

\newcommand{\Rmnum}[1]{\expandafter\@slowromancap\romannumeral #1@}
\makeatother
\makeatletter
\let\NAT@parse\undefined
\makeatother
\usepackage[pagebackref=false,breaklinks=true,colorlinks=true,bookmarks=false,linkcolor={red!50!black},urlcolor={blue},citecolor={green!60!black}]{hyperref} 

\usepackage{url}
\usepackage{pifont}
\usepackage{booktabs}
\usepackage{etoolbox}
\usepackage{float}

\makeatletter
\patchcmd{\@makecaption}
  {\scshape}
  {}
  {}
  {}
\makeatletter
\patchcmd{\@makecaption}
  {\\}
  {.\ }
  {}
  {}
\makeatother
\usepackage{pifont}
\usepackage{rotating}
\usepackage{amssymb}
\usepackage{amsmath}
\usepackage{CJKutf8}
\usepackage{booktabs}
\usepackage{multirow} 
\usepackage{mathrsfs}
\usepackage{graphicx}
\usepackage{etoolbox}
\usepackage{caption}
\usepackage{color}
\usepackage[table,dvipsnames]{xcolor}

\newcommand{\best}{\cellcolor{red!50}}
\newcommand{\sbest}{\cellcolor{orange!80}}
\newcommand{\tbest}{\cellcolor{yellow!80}}

\definecolor{my_green}{rgb}{0.85,0.97,0.74}
\definecolor{my_pink}{rgb}{0.996,0.835,0.906}
\definecolor{my_red}{rgb}{0.996,0.8,0.776}
\definecolor{my_yellow}{rgb}{0.984,0.843,0.522}
\definecolor{my_blue}{rgb}{0.725,0.906,0.996}
\definecolor{my_gt}{rgb}{0.996,0.843,0.745}

\title{\LARGE \bf
VDG: Vision-Only Dynamic Gaussian for Driving Simulation
}

\author{Hao Li $^{1,2,*}$, Jingfeng Li$^{1,*}$, Dingwen Zhang$^{1, \dagger}$, Chenming Wu$^2$, Jieqi Shi$^3$, Chen Zhao$^2$,\\ Haocheng Feng$^2$, Errui Ding$^2$, Jingdong Wang$^2$~\IEEEmembership{Fellow, IEEE}, Junwei Han$^1$~\IEEEmembership{Fellow, IEEE} 
\thanks{*This work was not supported by any organization}
\thanks{$^{1}$The School of Automation, Northwestern Polytechnical University
        {\tt\small albert.author@papercept.net}}%
\thanks{$^{2}$Department of Computer Vision Technology (VIS), Baidu Inc., {\tt\small albert.author@baidu.com}}
\thanks{$^{3}$HKUST Aerial Robotics Group, Hong Kong University of Science and Technology, HKSAR, China.
        {\tt\small b.d.researcher@ieee.org}}%
}

\begin{document}

\twocolumn[{
\renewcommand\twocolumn[1][]{#1}
\maketitle
\thispagestyle{empty}
\pagestyle{empty}
\begin{center}
    \captionsetup{type=figure}
    \vspace{-28pt} \includegraphics[width=\linewidth]{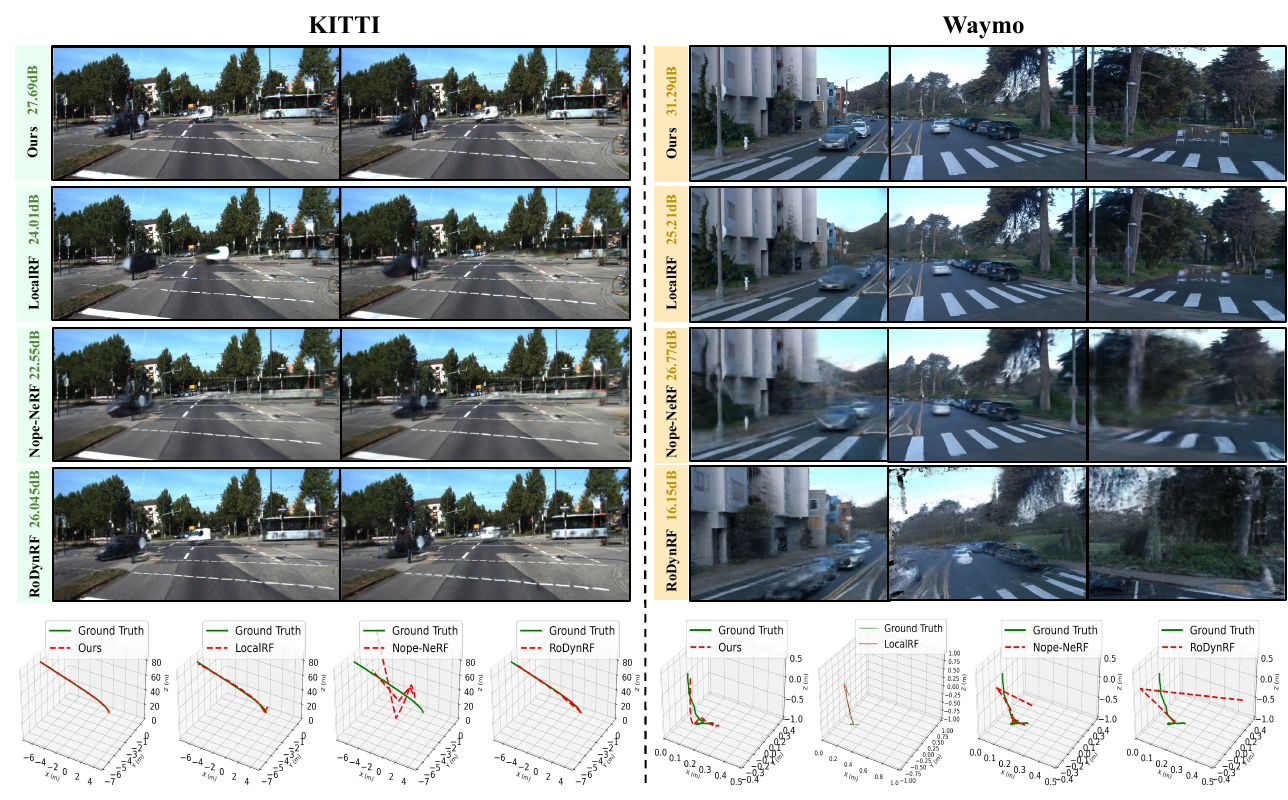}
    \captionof{figure}{Our proposed \textbf{VDG} is crafted to effectively and uniformly reconstruct large, dynamic urban scenes as well as predicted poses with only image input. Here, we showcase our reconstruction results and pose evaluation on KITTI~\cite{geiger2012we} and Waymo~\cite{waymo_open_dataset} datasets, and further compared with the latest pose-free methods. The reconstructed visualizations reveal that our method enables us to model static and dynamic objects without pose priors. Moreover, our method achieves much more accurate pose prediction than other pose-free methods.}
    \label{fig:overview}
\end{center}
}]
{
  \footnotetext[1]{BRAIN Lab, NWPU, China. {\tt\small lifugan\_10027@outlook.com  \{li0jingfeng, zhangdingwen2006yyy, junweihan2010\}@gmail.com}}
  \footnotetext[2]{Department of Computer Vision Technology (VIS), Baidu Inc., China. {\tt\small \{wuchenming, zhaochen02, fenghaocheng, dingerrui, wangjingdong\}@baidu.com}}
  \footnotetext[3]{Aerial Robotics Group, HKUST, Hong Kong, China. {\tt\small jshias@connect.ust.hk}}
  \renewcommand{\thefootnote}{\fnsymbol{footnote}}
  \footnotetext[1]{Equal Contribution} \footnotetext[2]{Corresponding Author}
}

\begin{abstract}

Dynamic Gaussian splatting has led to impressive scene reconstruction and image synthesis advances in novel views. 
Existing methods, however, heavily rely on pre-computed poses and Gaussian initialization by Structure from Motion (SfM) algorithms or expensive sensors. 
For the first time, this paper addresses this issue by integrating self-supervised VO into our pose-free dynamic Gaussian method (VDG) to boost pose and depth initialization and static-dynamic decomposition.
Moreover, VDG can work with RGB image input only and construct dynamic scenes at a faster speed and larger scenes compared with the pose-free dynamic view-synthesis method. 
We demonstrate the robustness of our approach via extensive quantitative and qualitative experiments. 
Our results show favorable performance over the state-of-the-art dynamic view synthesis methods. Additional video and source code will be posted on our project page at \href{https://3d-aigc.github.io/VDG}{https://3d-aigc.github.io/VDG}.
\end{abstract}

\section{INTRODUCTION}
Autonomous driving (AD) holds tremendous promise and could potentially revolutionize the future of transportation. Ensuring the safety of autonomous driving systems has become a primary focus of extensive development efforts. Driving simulation has gained significant prominence as a reliable, secure, and efficient alternative for training and evaluating AD software and algorithms~\cite{li2019aads,amini2020learning, amini2022vista, mars}. Simulation provides a robust platform that enables comprehensive testing and assessment in a controlled virtual environment, which is also a safe and controlled space for training, allowing for extensive testing of software and algorithms under various scenarios. A significant challenge in scalable driving simulation is to model and represent large-scale urban scenes~\cite{geiger2012we, li2023dgnr, li2024ho}.

Recently invented Neural Radiance Fields (NeRF) \cite{mildenhall2021nerf} and 3D Gaussian Splatting (3D-GS) \cite{chng2022gaussian} bridge the gap between computer vision and computer graphics in the tasks of image-based novel view synthesis and 3D reconstruction. By incorporating these synthetic views, it becomes feasible to expose the algorithms to a broader range of viewing conditions, thereby enhancing their robustness and generalization capabilities. This approach has the potential to significantly improve the performance of perception algorithms in real-world scenarios, contributing to the advancement of autonomous driving technology.
Approaches like MARS~\cite{mars}, PNF~\cite{kundu2022panoptic}, and Driving-Gaussian~\cite{zhou2023drivinggaussian} decompose scenes into objects and backgrounds by integrating detection or segmentation auxiliary tasks. However, obtaining accurate object-level supervision in the real world is challenging, and manual annotations are always expensive. Other approaches (PVG~\cite{chen2023periodic}, EmerNeRF~\cite{yang2023emernerf}, \textit{etc.}) decompose dynamic and static scenes in a self-supervision manner, while they still require additional LiDAR point clouds and accurate pose priors as input. 

Besides, an intuitive approach involves pre-processing the scene using Structure from Motion (SfM) such as COLMAP~\cite{schonberger2016structure}. Although it can get poses and coarse point clouds for Gaussian initialization, it still suffers from long-time consumption. To address this issue, several NeRF-based~\cite{bian2023nope, lin2021barf, yen2021inerf, wang2021nerf, fu2023mononerf} and 3DGS-based~\cite{fu2023colmap, keetha2023splatam, hu2024cg} methods have been developed. By introducing a state-of-the-art depth estimation network~\cite{deng2022depth} or RGB-D cameras and learnable poses, these methods can train NeRF / Gaussian parameters and recover camera poses without requiring camera poses. However, these methods are only confined to static scenes on a small scale and fail to handle forward-moving driving scenarios. 
RoDynRF~\cite{liu2023robust} optimizes NeRF, and camera poses in dynamic scenes. However, it requires additional computation for forward/backward flow, leading to a significant workload, even with just a dozen images, thus making it difficult to apply in scalable driving simulations.

%

%

%

%

%

%
%
%
%

Our work aims to tackle the aforementioned issues by proposing the Vision-only Dynamic Gaussian (VDG), which can construct entire \textbf{dynamic} driving scenes represented by 3DGS~\cite{kerbl3Dgaussians} with only vision input (\textit{i.e.}, \textbf{pose-free}).
In particular, our method introduces a self-supervised visual odometry approach that yields precise camera pose estimation. Leveraging its self-supervised nature, our method additionally provides dense depth estimation, which is on par with the scale of pose estimation. This dense depth estimation plays a crucial role in Gaussian initialization, enhancing the effectiveness of our approach.
Furthermore, we propose a scene decomposition step in our framework that separates scenes into static and dynamic components. To achieve improved results, we parameterize and update the poses accordingly. Since our framework employs a self-decomposing approach, decomposing scenes without ground-truth poses becomes challenging. To address this, we leverage motion masks generated by the visual odometry and introduce a motion-mask supervised mechanism. This mechanism boosts the network's ability to identify dynamic objects within the scenes.
We summarize the contributions of this paper as follows. 
\begin{itemize}
    \item We propose a novel method, VDG, that introduces self-supervised visual odometry for camera pose and dense depth estimation, enabling training dynamic Gaussian scenes for driving simulation with vision-only inputs without pose priors.
    \item We further incorporate a motion-mask supervised mechanism to decompose scenes into static and dynamic components, improving the network's ability to identify dynamic objects.
    \item We conduct comprehensive experiments on urban scene datasets (\textit{i.e.} Waymo and KITTI), demonstrating that our VDG method surpasses state-of-the-art baselines and delivers superior quantitative and qualitative results.
\end{itemize}

\section{Related Work}
\subsection{Novel View Synthesis for Driving Simulation}
Recently, there have been significant advancements in scene reconstruction and novel view synthesis, thanks to advanced implicit neural techniques~\cite{mildenhall2021nerf} and explicit representation methods~\cite{chng2022gaussian}. These techniques have played a pivotal role in various applications, including autonomous driving~\cite{waymo_open_dataset}, auto navigation~\cite{Kwon_2023_CVPR}, high-fidelity data generation~\cite{wu2023reconfusion}, and digital maps~\cite{caesar2020nuscenes}.
Simultaneously, it is important to acknowledge that our world exhibits inherent dynamic and complex characteristics in both spatial and temporal dimensions, particularly in urban scenes.
NeRF-based techniques, such as \cite{pumarola2021d,park2021hypernerf,wu2022d,yang2023emernerf,martin2021nerf}, have extended NeRF's static scene assumption to dynamic environments. However, these methods face various challenges, such as handling large-scale scenes, long training times, and the absence of geometric priors regarding temporal influences on scenes.

More recently, 3D Gaussian Splatting (3D-GS)~\cite{kerbl3Dgaussians} has introduced innovative explicit 3D scene representation variants allowing fast training and real-time rendering. Building upon these variants, Driving-Gaussian \cite{zhou2023drivinggaussian} and PVG \cite{chen2023periodic} facilitate the decomposition of static and dynamic Gaussians by incorporating LiDAR priors. Additionally, Street-Gaussians \cite{yan2024street} introduces 3D object detection to identify and decompose dynamic objects within scenes.
However, the above methods require additional inputs, such as precise camera parameters and expensive LiDAR hardware. In contrast, our method achieves comparable performance in reconstructing urban scenes using only collected images as input.

\subsection{Pose-free Novel View Synthesis}
Accurate camera poses are essential for both NeRF-based and 3DGS-based methods in novel view synthesis and image reconstruction. The initial attempt towards pose-free novel view synthesis was iNeRF \cite{yen2021inerf}, which estimated poses by matching keypoints using a pre-trained neural radiance field model. Another method, Nope-NeRF \cite{bian2023nope}, improved pose estimation by incorporating monocular depth estimation to constrain relative poses between images.
However, these NeRF-based methods struggle with large-scale scenes due to inaccurate poses. To address this limitation, \(\mathrm{F}^2\)-NeRF and LocalRF split the scene into multiple blocks, utilizing several NeRF models to reconstruct the entire scene. Nonetheless, these methods still suffer from slow training speed and lack geometric priors.

More recently, various works \cite{Yu2023MipSplatting, chng2022gaussian} have leveraged Structure-from-Motion (SfM) \cite{schonberger2016structure} to solve camera poses and initialize Gaussian points. Unfortunately, this preprocessing step incurs additional computational burden. Therefore, several pose-free 3D-GS methods, such as SplaTAM \cite{keetha2023splatam}, CF-3DGS \cite{fu2023colmap}, GS-SLAM \cite{yan2023gs}, and others, have introduced RGBD or off-the-shelf depth estimation networks for Gaussian initialization and parameterized poses for pose optimization. However, these strategies face challenges in reconstructing large-scale urban scenes, particularly when dynamic objects disrupt the pose regression. Furthermore, handling large scenes and free camera trajectories remains challenging for learnable poses without scene decomposition.
In contrast to previous methods, our VDG approach enables the reconstruction of entire urban scenes and conducts novel view synthesis using only input images.

\section{Preliminary}
3D Gaussian Splatting (3D-GS)~\cite{kerbl3Dgaussians} represents a 3D scene as a collection of 3D Gaussian primitives $\mathbb{G} = \{ (\mu_k, \Sigma_k, \alpha_k, S_k)\}^K$. Each Gaussian primitive consists of a mean $\mu_k$, a covariance $\Sigma_k$, an opacity $\alpha_k$, and spherical harmonics coefficients $S_k$. These primitives parameterize the 3D radiance field of the scene through a tiled-based rasterization process; 3DGS facilitates real-time alpha blending of numerous Gaussians to render novel-view images. 

The 3D primitives are projected to 2D screen space and rasterized using $\alpha$-blending. The $\alpha$-blending weights are given as: \(\alpha = oG\) with the projected Gaussian contribution on pixel $(x, y)$ given as:
\begin{equation}
    G(x, y) = e^{-\frac{1}{2} ([x, y]^T - \mu')^T \Sigma'^{-1} ([x, y]^T - \mu')} 
\end{equation}
where $\mu'$ and $\Sigma'$ are the projected 2D mean and covariance matrix, respectively.
The combined effect of converting SHs to per-view color values and $\alpha$-blending them recreates the appearance of the captured scene.

\section{Proposed Method}
\begin{figure*}
    \centering
    \includegraphics[width=1\linewidth]{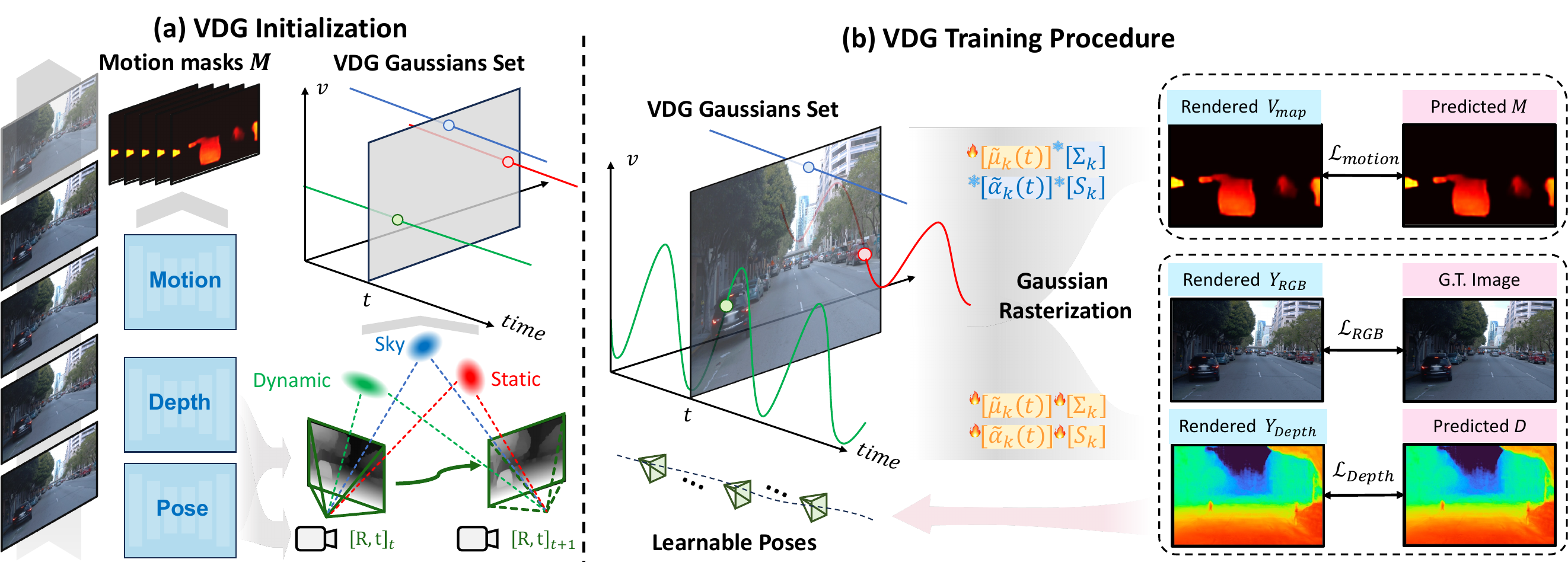}
    \captionof{figure}{The proposed VDG. (a) VDG Initialization: uses the off-the-shelf VO network \(\mathcal{P}(\cdot)\), \(\mathcal{M}(\cdot)\), and \(\mathcal{D}(\cdot)\) to estimate the global poses \(T_t\), motion masks \(M_t\), and depth maps \(D_t\) (see Sec. \ref{sec:vo}). Given poses \(T_t\) and corresponding depth maps \(D_t\), we project the depth maps into 3D space to initialize the Gaussian points \(G^k_t =\{\tilde{\mu}^k_t, \Sigma^k, \widetilde{\alpha}^k_t, S^k\}\). Note that the velocity \(v\) of each Gaussian is set to 0 (see Sec. \ref{sec:init}). (b) VDG Training Procedure: Given initialized Gaussians \(G^k_t\), we train our VDG using RGB and depth supervision (see Sec. \ref{sec:train}). Moreover, we apply motion mask supervision to decompose static and dynamic scenes (Sec. \ref{sec:motion}). In the end, we adopt a training strategy to refine vo-given poses \(T_t\) (Sec. \ref{sec:strategy}).}
    \label{fig:pipeline}
    \vspace{-10pt}
\end{figure*}

Given an arbitrary sequence of unposed images alongside their corresponding timestamps \(\left \{ (\mathit{I}_i,t_i)  \mid i=1, \cdots, N\right\}\) in an unbounded scene(\textit{i.e.} Autonomous Driving), VDG aims to achieve precise 3D reconstruction and synthesize novel view at any desired timestamp \(t\) and camera pose \([\mathrm{R}, \mathrm{t}]\).
To accomplish the above objective, we develop VDG that leverages self-supervised VO for accurate pose and monocular depth estimation for our Gaussian initialization (Sec. \ref{sec:overview}). Besides, motion supervision is proposed to decompose dynamic and static scenes for better reconstruction (Sec. \ref{sec:motion}). 
In the end, the training strategy and optimization for large-scale scenarios are introduced in sec. \ref{sec:strategy}.
\subsection{Overall Framework}
\label{sec:overview}
%
\subsubsection{Off-the-Shelf VO Inference}
\label{sec:vo}
Given the advantage that self-supervised VO uses a photometric mechanism to generate depth and pose,  the predicted depth and pose are aligned in the same scale, which is necessary for Gaussian initialization. 
Therefore, with two adjacent images $\mathit{I}_t,\mathit{I}_{t-1}$, the self-supervised VO \cite{sun2024dynamo} are used to estimate their relative pose \(T_{t\rightarrow t-1}\), motion mask \(M_{t\rightarrow t-1}\) and \(I_t\)'s depth map \(D_t\) via ego-motion pose network \(\mathcal{P}\), depth network \(\mathcal{D}\) and motion mask \(\mathcal{M}\):
\begin{equation}
    \begin{aligned}
    T_{t\rightarrow t-1} & =\mathcal{P} (I_{t},I_{t-1}) , \quad & T_{t\rightarrow t-1} & \in \mathbb{R}^{3\times 4} \\ 
    M_{t\rightarrow t-1}& =\mathcal{M}(I_{t},I_{t-1}), \quad  & M_{t\rightarrow t-1} & \in \mathbb{R}^{H\times W} \\
    D_{t} & =\mathcal{D} (I_{t}), \quad &  D_{t} & \in \mathbb{R}^{H\times W}  
    \end{aligned} 
\end{equation}
With relative pose \(T_{t\rightarrow t-1}\) from the previous time, one might easily infer the absolute pose \([\mathrm{R}_t, \mathrm{t}_t]\). 
\subsubsection{Gaussian Initialization on Single View}
\label{sec:init}
After that, for each time step \(t\), we use the predicted depth map \(D_t\) and the corresponding absolute pose \([\mathrm{R}_t, \mathrm{t}_t]\) to initialize 3DGS \(\mathbb{G}_t\) with points lifted from monocular depth, leveraging camera intrinsic and orthogonal projection, instead of the widely used SfM points.
\subsubsection{Network Training}
\label{sec:train}
With the initialized Gaussian points \(\mathbb{G}_t\), pose \([\mathrm{R}_t, \mathrm{t}_t]\), and timestamp \(t\), we then train our dynamic Gaussian and further optimize our pose given the fact that the process is differentiable. We adopt a motion supervision mechanism during the training to decompose static and dynamic Gaussians better using motion mask \(M_{t\rightarrow t-1}\). Ultimately, a training strategy is further proposed to maintain our geometry representation during pose optimization.

\subsection{Static-Dynamic Scenes Model and Decomposition}
\label{sec:motion}
Traditional 3D-Gaussian cannot model time-varying scenes like urban scenes, which possess many dynamic objects
Therefore, following the idea of PVG~\cite{chen2023periodic}, our proposed VDG decomposes static background and dynamic objects in a self-supervision manner. It modifies the traditional Gaussian's mean $\mu_k$ and opacity $\alpha_k$ to be time-dependent functions centered around the life peak $\tau$, denoted as $\tilde{\mu}(t)$ and $\tilde{\alpha}(t)$.  Formally, each Gaussian \(G_k(t)\) can be expressed as:
\begin{equation}
\label{eq:1}
\begin{aligned}
G^k_t & =\{\tilde{\mu}^k_t, \Sigma^k, \widetilde{\alpha}^k_t, S^k\}, \\
\widetilde{\mu}^k_t & =\mu^k+\frac{l}{2 \pi} \cdot \sin \left(\frac{2 \pi(t-\tau)}{l}\right) \cdot v^k, \\
\widetilde{\alpha}^k_t & =\alpha^k \cdot e^{-\frac{1}{2}(t-\tau)^2 \beta^{-2}},
\end{aligned}
\end{equation}
where hyper-parameter \(l\) represents the cycle length, serving as the scene prior, learnable parameters \(v_k=\left.\frac{\mathrm{d} \tilde{\mu}(t)}{\mathrm{d} t}\right|_{t=\tau} \in \mathbb{R}^3\) denotes the instant velocity at time \(\tau\). 

\begin{figure}[b]
    \centering
    \includegraphics[width=\linewidth]{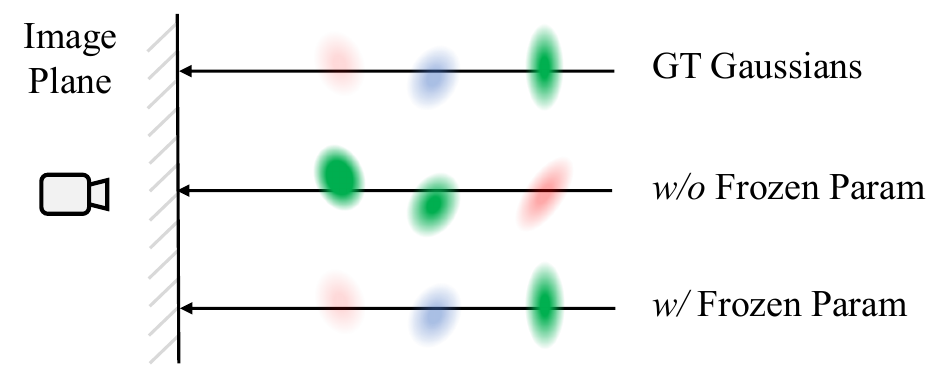}
    \caption{Illustration of Frozen Gaussian Parameters. }
    \label{fig:frozen}
\end{figure}

However, unlike pose-given methods like PVG, decomposing dynamic objects and estimating our poses simultaneously can be challenging. 
It is caused by two factors: 
(1) changes in observation can be attributed to either camera motion or the Gaussians' ability to model view-dependent appearances, which disturbs the parameterized pose optimization with photometric supervision; 
(2) The estimated poses provided by VO are still rough.
To this end, we need a more explicit way to guide our VDG in decomposing static and dynamic scenes. 
Intuitively, we propose motion mask supervision \(\mathcal{L}_{motion}\) based on the motion masks \(M \in \{0,1\}\) produced by VO. 
Given the rendered velocity map \(V\), instead of directly using L2 loss, we convert the smooth velocity \(v_k \in [0,1]\) from Eq. \ref{eq:1} to a discrete velocity \(\hat{v}_k \in \{0,1\}\) before the L2 loss. The formation is shown below:
\begin{equation}
\begin{aligned}
& \hat{V}=  V + \not \nabla \left[ \left(V>v_{thr}\right)-V\right] \\
& \mathcal{L}_{motion}  =  ||\hat{V} - M ||^2_2,
\end{aligned}
\end{equation}
where \(v_{thr}\) is the hyper-parameters that threshold the static Gaussians and dynamic Gaussians, and \(\not \nabla[\cdot]\) is the stop-gradient operator.
It is essential to make diversity between the static and dynamic Gaussians while keeping the proper distribution of dynamic Gaussians on different \(v\). 
Notably, as shown in Fig. \ref{fig:frozen}, directly adopting \(\mathcal{L}_{motion}\) could lead to a negative impact on the final render quality. Because the network tends to overfit the dynamic supervision by changing the parameters of the Gaussian point (\textit{e.g.} location, opacity, and colors).  

\subsection{Training Strategy}
In the early stage of the training, we use \([\mathrm{R}_t, \mathrm{t}_t]\) as a pose before training our Gaussian model.
However, imperfect poses will limit the final render quality of our Gaussian model.
Therefore, at the end of the training phase, we adopt pose refinement by translating poses \([\mathrm{R}_t, \mathrm{t}_t]\) into learnable parameters \([\tilde{\mathrm{R}}_t, \tilde{\mathrm{t}}_t]\) to achieve more precise scene reconstruction, as shown in Fig. \ref{fig:overview}. 
Moreover, to eliminate the negative impact of moving objects for pose optimization, we only use static parts (\textit{e.g.} Gaussians with low velocity \(\hat{V}\)) to optimize our refined poses.

\label{sec:strategy}

\begin{figure}
\centering
    \includegraphics[width=\linewidth]{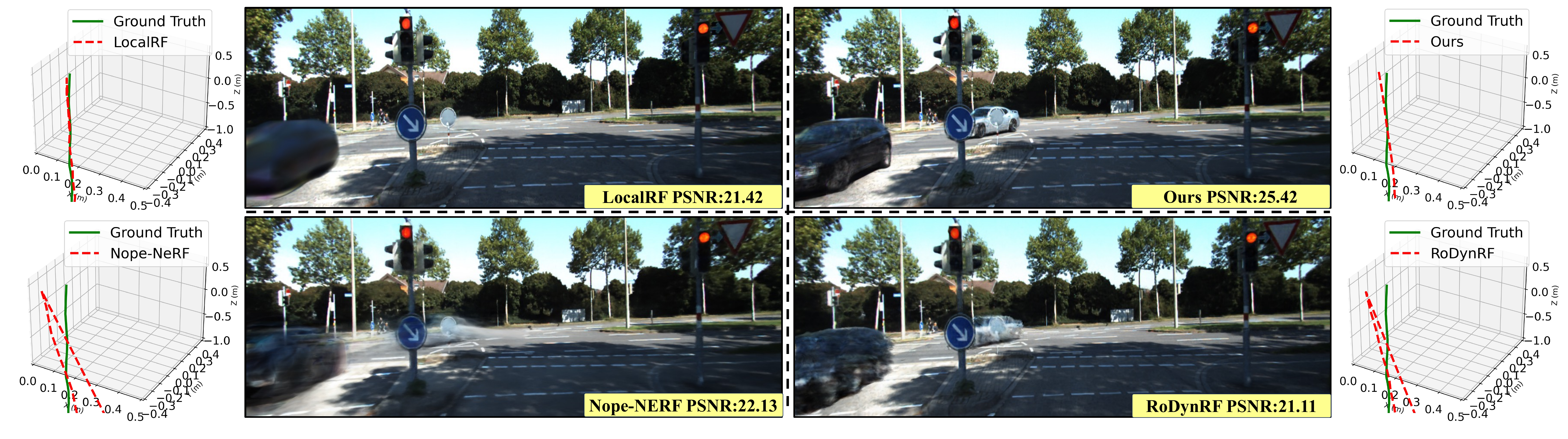}
    \captionof{figure}{Qualitative comparison on KITTI dataset regarding pose accuracy and rendered quality. Our method outperforms other baselines, even in cases where pose estimation is relatively poor.}
    \label{fig:kitti_worse}
\end{figure}

\begin{table*}[!t]
\centering
\renewcommand{\arraystretch}{1.2} 
\setlength{\tabcolsep}{3pt} 
\caption{Quantitative performance of novel view synthesis on the Waymo Open Dataset~\cite{waymo_open_dataset} and KITTI benchmark~\cite{geiger2012we}. '-' means SplaTAM cannot rendering original resolution image on a single NVIDIA V100 GPU.}
\small
\begin{tabular}{cl|ccccccc|ccccccc}
\toprule
\multicolumn{2}{c|}{\multirow{3}{*}{Methods}} & \multicolumn{7}{c|}{Waymo Open Dataset~\cite{waymo_open_dataset}}                                                                                               & \multicolumn{7}{c}{KITTI Dataset~\cite{geiger2012we}}                                                                                                          \\
\multicolumn{2}{c|}{}                         & \multicolumn{1}{c|}{\multirow{2}{*}{FPS}} & \multicolumn{3}{c|}{Image   Reconstruction} & \multicolumn{3}{c|}{Novel View   Synthesis} & \multicolumn{1}{c|}{\multirow{2}{*}{FPS}} & \multicolumn{3}{c|}{Image   Reconstruction} & \multicolumn{3}{c}{Novel View Synthesis} \\
\multicolumn{2}{c|}{}                         & \multicolumn{1}{c|}{}                     & PSNR $\uparrow$  & SSIM$\uparrow$  & \multicolumn{1}{c|}{LPIPS$\downarrow$} & PSNR$\uparrow$          & SSIM$\uparrow$         & LPIPS$\downarrow$        & \multicolumn{1}{c|}{}                     & PSNR$\uparrow$   & SSIM$\uparrow$  & \multicolumn{1}{c|}{LPIPS$\downarrow$} & PSNR$\uparrow$         & SSIM$\uparrow$        & LPIPS$\downarrow$       \\ \midrule
\multirow{5}{*}{\begin{sideways}Pose-free \ding{56}\end{sideways}}   & StreetSurf   & \multicolumn{1}{c|}{0.097}                & 27.00  & 0.850 & \multicolumn{1}{c|}{0.361} & 21.83         & 0.776        & 0.416        & \multicolumn{1}{c|}{0.370}                 & 24.14  & 0.819 & \multicolumn{1}{c|}{0.257} & 22.48        & 0.768       & 0.304       \\
                               & SUDS         & \multicolumn{1}{c|}{0.008}                & 28.83  & 0.805 & \multicolumn{1}{c|}{0.289} & 21.83         & 0.656        & 0.405        & \multicolumn{1}{c|}{0.040}                 & 28.31  & 0.876 & \multicolumn{1}{c|}{0.185} & 22.77        & 0.797       & 0.171       \\
                               & EmerNeRF     & \multicolumn{1}{c|}{0.053}                     &        28.11&       0.786& \multicolumn{1}{c|}{0.373}      &               25.92&              0.763&              0.384& \multicolumn{1}{c|}{0.280}                     &        26.95&       0.828& \multicolumn{1}{c|}{0.218}      &              25.24&             0.801&             0.237\\
                               & 3D-GS        & \multicolumn{1}{c|}{63.00}                   & 27.99  & 0.866 & \multicolumn{1}{c|}{0.293} & 25.08         & 0.822        & 0.319        & \multicolumn{1}{c|}{125.0}                  & 21.02  & 0.811 & \multicolumn{1}{c|}{0.202} & 19.54        & 0.776       & 0.224       \\
                               & PVG          & \multicolumn{1}{c|}{50.00}                   & \textbf{32.46}  & \textbf{0.910} & \multicolumn{1}{c|}{0.229} & \textbf{28.11}         & \textbf{0.849}        & \textbf{0.279}        & \multicolumn{1}{c|}{59.00}                   & 31.54  & 0.927 & \multicolumn{1}{c|}{0.083} & \textbf{26.63}        & \textbf{0.885}       & \textbf{0.127}       \\ \midrule
\multirow{5}{*}{\begin{sideways}Pose-free \ding{52}\end{sideways}}   & Nope-NeRF    & \multicolumn{1}{c|}{0.516} & \sbest 25.26 & \tbest 0.756& \multicolumn{1}{c|}{\tbest 0.536}      & \tbest 21.75& \tbest 0.738& \tbest 0.547& \multicolumn{1}{c|}{0.518} & \sbest 23.54& \tbest 0.756& \multicolumn{1}{c|}{ \tbest 0.325}      & \tbest 22.54& \tbest 0.747& \tbest 0.327\\
                               & LocalRF      & \multicolumn{1}{c|}{0.046}                 &   \tbest   25.20 &    \sbest   0.800& \multicolumn{1}{c|}{ \sbest 0.409}      & \sbest  23.77& \sbest 0.775& \sbest 0.425& \multicolumn{1}{c|}{0.143}            & \tbest 22.56& \sbest 0.861& \multicolumn{1}{c|}{ \sbest 0.136}      & \sbest 23.48& \sbest 0.864& \sbest 0.135\\
                               & SplaTAM      & \multicolumn{1}{c|}{-}                     &        -&       -& \multicolumn{1}{c|}{-}      &               -&              -&              -& \multicolumn{1}{c|}{360.0}                     &        17.51&       0.716& \multicolumn{1}{c|}{0.291}      &              17.14&             0.691&0.313             \\
                               & RoDynRF       & \multicolumn{1}{c|}{0.010}                &        16.03&       0.617& \multicolumn{1}{c|}{0.615}      &               14.94&              0.595&              0.627& \multicolumn{1}{c|}{0.040}                     &        26.04&       0.817& \multicolumn{1}{c|}{0.239}      &              17.47&             0.544&             0.371\\
                               & Ours         & \multicolumn{1}{c|}{44.54}                 &   \best    31.65 &  \best    0.900 & \multicolumn{1}{c|}{ \best 0.250} &     \best 26.68 & \best 0.817 & \best 0.324 & \multicolumn{1}{c|}{60.86} &    \best \textbf{31.61}&   \best \textbf{0.933}    & \multicolumn{1}{c|}{\best \textbf{0.074}}      &            \best  25.29&     \best        0.851&         \best    0.152 \\ \bottomrule 
\label{view synthesis table}
\end{tabular}
\end{table*}

\begin{figure*}
    \vspace{-20pt}
    \centering
    \includegraphics[width=\linewidth]{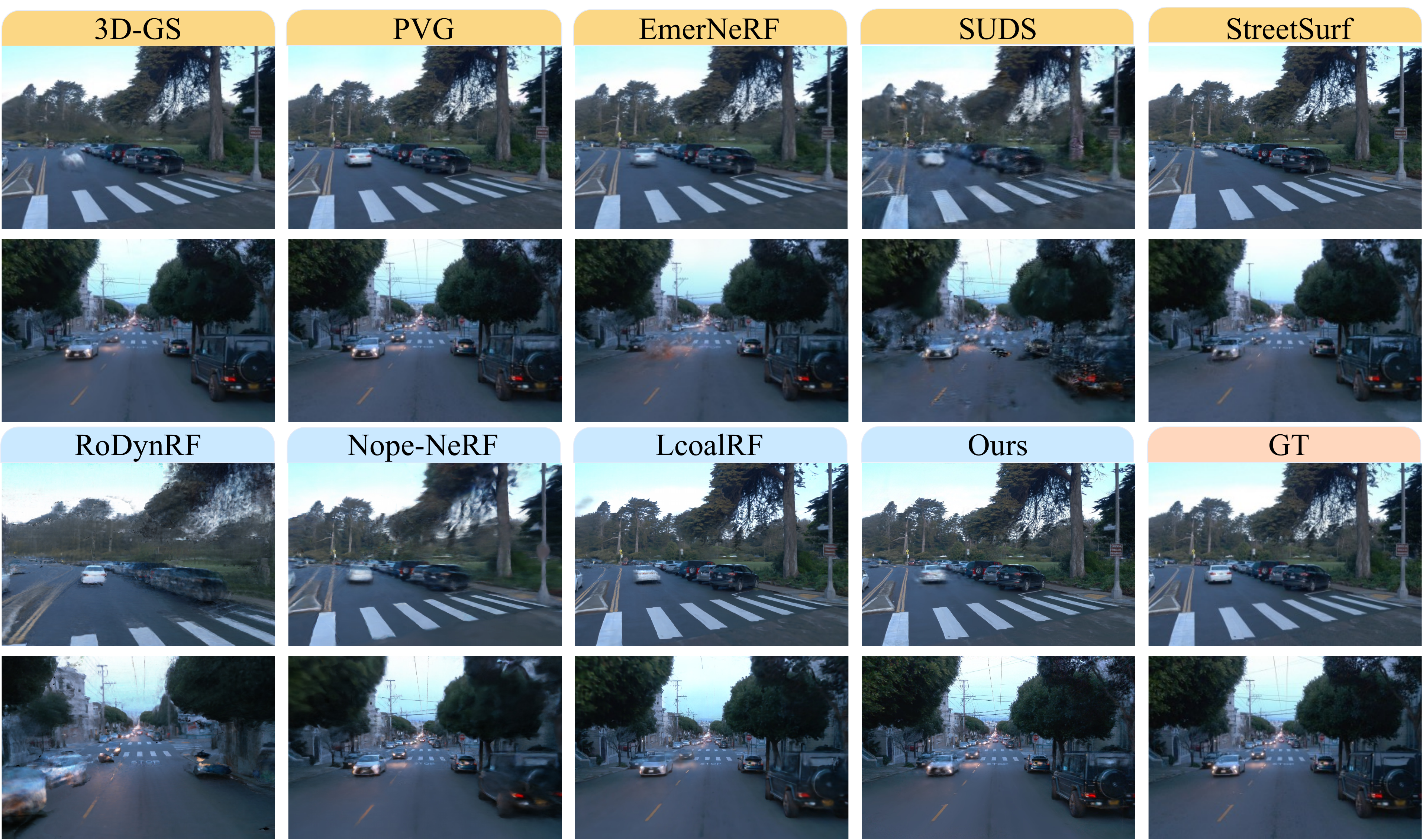}
    \captionof{figure}{\setlength{\fboxsep}{0pt}Qualitative comparison of our approach and other baselines, including \colorbox{my_blue}{Pose-free methods}, \colorbox{my_yellow}{GT pose needed methods} and \colorbox{my_gt}{GT images} on Waymo Open Dataset\cite{waymo_open_dataset}. We show two case views of synthesis results under different scenes.}
    \vspace{-0.6cm}
    \label{fig:waymo_results}
\end{figure*}
\section{EXPERIMENTS}
\subsection{Experimental Setup}
\noindent
\textbf{Datasets.} We conduct detailed experiments on two datasets, KITTI benchmark\cite{geiger2012we} and Waymo Open Dataset\cite{waymo_open_dataset}, with large amounts of moving objects, especially fast ego-camera motion, and complex light conditions. show the quantitative results in Tab.~\Rmnum{1}. We selected 4 dynamic scenes in Waymo and 3 dynamic scenes in KITTI to evaluate pose accuracy and novel view synthesis. We adopt a novel view for each scene by selecting every 4th frame in sequences. And we use original resolution for training and rendering.\\

\textbf{Baselines.} Our approach primarily surpasses pose-free methods and is comparable in performance to other methods with accurate poses. (a) pose-free methods: NoPe-NeRF\cite{bian2023nope}, LocalRF\cite{meuleman2023progressively}, RoDynRF\cite{liu2023robust}, SplaTAM\cite{keetha2023splatam}. (b) other methods with GT poses: EmerNeRF\cite{yang2023emernerf}, StreetSurf\cite{guo2023streetsurf}, SUDS\cite{turki2023suds}, 3D-GS\cite{kerbl3Dgaussians}, PVG\cite{chen2023periodic}. Furthermore, we evaluate image reconstruction, novel view synthesis, and estimated pose accuracy tasks. Due to the length of AR dataset sequences and the forward motion of cameras in autonomous driving scenes, COLMAP often struggles to localize.\\
\noindent
\begin{figure*}
\centering
    \includegraphics[width=\linewidth]{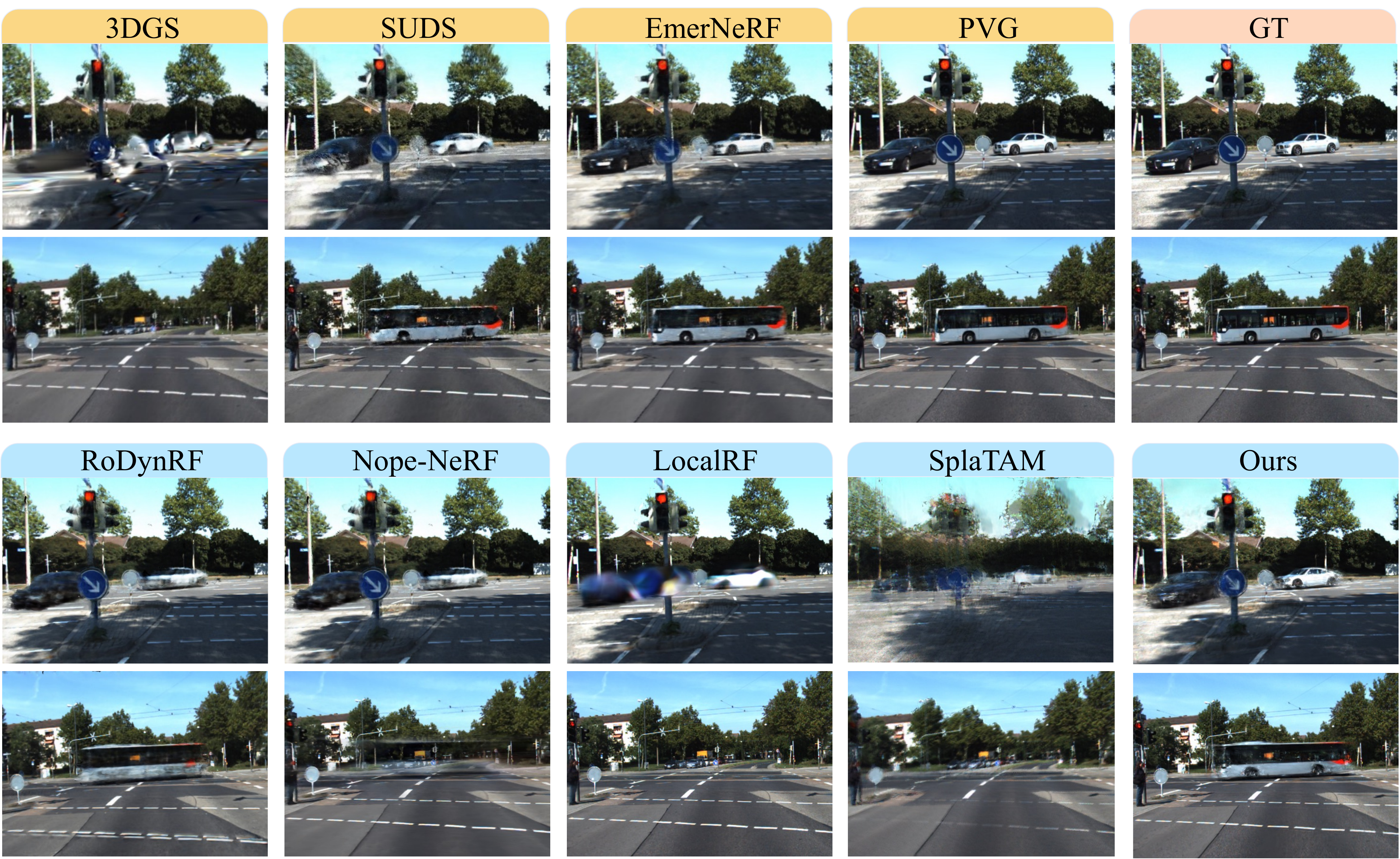}
    \captionof{figure}{\setlength{\fboxsep}{0pt}Qualitative comparison of our approach and other baselines, including \colorbox{my_blue}{Pose-free methods}, \colorbox{my_yellow}{GT pose needed methods} and \colorbox{my_gt}{GT images} on KITTI benchmark\cite{geiger2012we}. We show two case views of synthesis results under different scenes.}
    \label{fig:kitti_results}
\end{figure*}
\textbf{Metrics.} We follow the evaluation protocol like NoPe-NeRF. We evaluate the model using two primary tasks: novel view synthesis and camera pose estimation. For novel view synthesis, we use Peak Signal-to-Noise Ratio (PSNR), Structural Similarity Index Measure (SSIM), and Learned Perceptual Image Patch Similarity (LPIPS). We evaluate camera pose accuracy using standard evaluation metrics, including the absolute trajectory error (ATE) and relative pose error (RPE). ATE measures the absolute difference between predicted camera positions and ground truth positions. RPE measures relative rotation and translation errors between consecutive images. The estimated trajectory is aligned with ground truth using Sim(3) with 7 DoF. \\
\noindent
\textbf{Implementation Details.} We initialize Gaussian points exclusively using the point cloud derived from the estimated monocular depth map projection. The color of the point cloud is acquired by mapping the points onto the corresponding image plane and retrieving the corresponding pixel values. Our train process involves 30k/40k iterations with the Adam optimizer\cite{kingma2017adam} that requires around 2 hours. All experiments are conducted on a single NVIDIA V100 GPU.

\subsection{Evaluation on Dynamic View Synthesis}
\begin{figure*}
\centering
    \includegraphics[width=\linewidth]{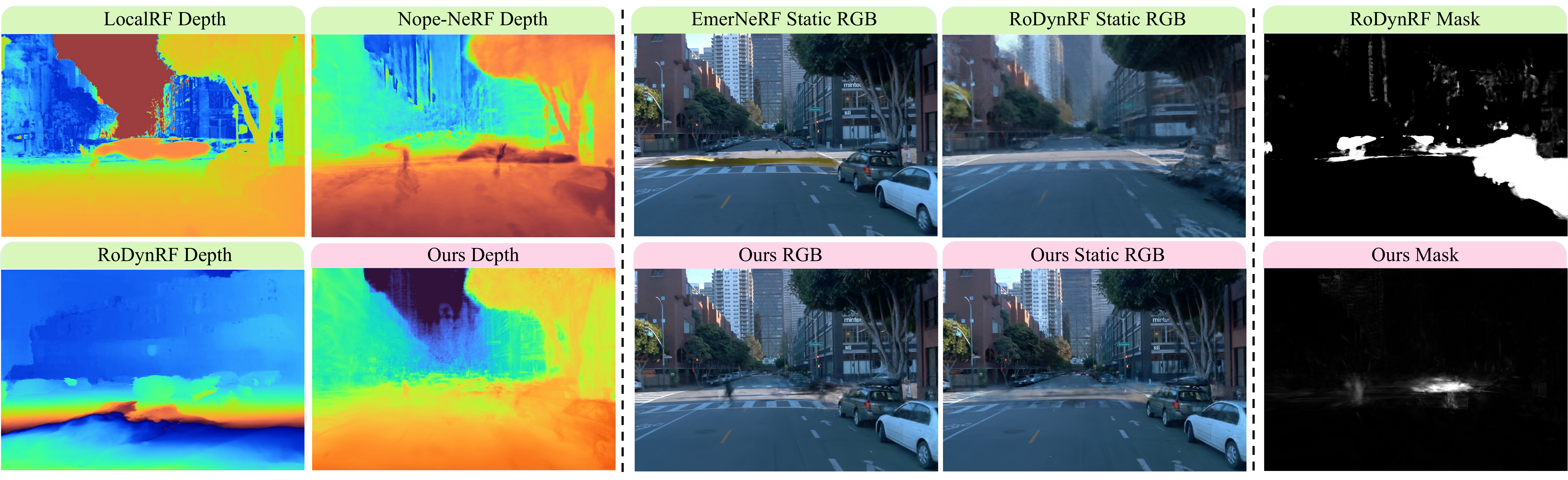}
    \captionof{figure}{\setlength{\fboxsep}{0pt}Qualitative comparisons of \colorbox{my_green}{other baselines} and \colorbox{my_pink}{our method} in terms of RGB, depth, binary mask, and static RGB.}
    \label{fig:results}
\end{figure*}

\begin{table*}[!t]
\centering
\renewcommand{\arraystretch}{1.2} 
\setlength{\tabcolsep}{3pt} 
\caption{Pose accuracy on the Waymo Open~\cite{waymo_open_dataset} and KITTI~\cite{geiger2012we} datasets. Note that the unit of $RPE_r$ is in degrees, ATE is in the ground truth scale and $RPE_t$ is scaled by 100.}
\small
\begin{tabular}{cccccccccccccccccc}
\hline
 & \multicolumn{1}{r}{\multirow{2}{*}{scenes}} &  & \multicolumn{3}{c}{Ours}              &  & \multicolumn{3}{c}{Nope-NeRF}         &  & \multicolumn{3}{c}{RoDynRF}           &  & \multicolumn{3}{c}{LocalRF}           \\ \cline{4-6} \cline{8-10} \cline{12-14} \cline{16-18} 
 & \multicolumn{1}{r}{} &  & \multicolumn{1}{c}{$RPE_t\downarrow$} & $RPE_r\downarrow$ & $ATE\downarrow$ &  & \multicolumn{1}{c}{$RPE_t\downarrow$} & $RPE_r\downarrow$ & $ATE\downarrow$ &  & \multicolumn{1}{c}{$RPE_t\downarrow$} & $RPE_r\downarrow$ & $ATE\downarrow$ &  & \multicolumn{1}{c}{$RPE_t\downarrow$} & $RPE_r\downarrow$ & $ATE\downarrow$ \\ \hline
\multirow{5}{*}{\rotatebox{90}{Waymo}}  &0017085&  &   2.619&      0.049&     0.112&  &    271.658&      1.921&     8.366&  &   235.341&      1.799&     2.400&  &   205.941&      1.745&     0.898\\
 & 0145050 &  &   4.563&      0.059&     0.175&  &   208.550&      1.760&     8.885&  &    217.774&      0.240&     2.580&  &   185.714&      0.253&     0.567\\
 &   0147030&  &   2.040&      0.031&     0.069&  &   206.084&      0.222&     7.041&  &   200.030&      0.134&     2.465&  &   169.140&      0.132&     0.539\\
 &   0158150&  &   1.829&      0.039&     0.080&  &   178.742&      0.536&     3.417&  &   254.535&      1.826&     5.380&  &   139.159&      0.620&     0.450\\
 &   Avg.&  &  \best 2.763& \best 0.045& \best 0.109&  &  \tbest 216.259&      1.110&     6.941&  &   235.999&    \best  1.030&   \tbest  4.698&  &  \sbest 174.989& \sbest 0.688& \sbest  0.614\\ \hline
\multicolumn{1}{l}{\multirow{4}{*}{\rotatebox{90}{KITTI}}} &   0001&  &   16.723&      0.152&     0.074&  &   9.501&      0.078&     0.457&  &   10.338&      0.066&     0.469&  &   5.590&      0.047&     0.201\\
\multicolumn{1}{l}{}&   0002&  &   4.536&      0.059&     0.175&  &   3.251&      0.038&     0.194&  &   3.335&      0.017&     0.198&  &   1.683&      0.019&     0.059\\
\multicolumn{1}{l}{}&   0006&  &   0.127&      0.018&     0.006&  &   0.162&      0.078&     0.007&  &   0.133&      0.026&     0.007&  &   0.121&      0.018&     0.006\\
\multicolumn{1}{l}{}&   Avg.&  &   7.262&      0.069&  \best   0.058&  &  \tbest 4.305&   \tbest   0.065&     0.219&  &  \sbest 4.602&    \sbest  0.036&     0.225&  &  \best 2.465&     \best 0.028&  \tbest   0.089\\ \hline
\end{tabular}
\label{pose tab1}
\vspace{-0.4cm}
\end{table*}

In this section, we primarily analyze the dynamic and static components decoupling effects of our method and the images synthesized through our approach are significantly clearer than those produced by other pose-free methods. As illustrated in Fig.~\ref{fig:results}, The first two columns show the comparative results of geometric in novel view, and the last three columns demonstrate the ability of our method to distinguish between dynamic Gaussian points and static Gaussian points. 

Unlike other w/ GT pose methods in which the camera poses of testing views are given, pose-free methods freeze the pre-trained NeRF/Gaussian Splatting model that trained on the training views and optimize the testing view's camera poses via training loss. Table\ref{view synthesis table} presents the comparison results of our approach with other baselines\cite{bian2023nope,liu2023robust,meuleman2023progressively} on the KITTI and Waymo dataset. Our method significantly outperforms other baselines in all metrics, including rendering speed and rendering quality. 
Notably, NeRF-based methods take significantly longer training time than our approach, and their rendering results are much worse than ours. Our method achieves comparable performance with well-known methods, even 3DGS and SUDS, which rely on LiDAR point cloud. In this subsection, we show some visualization results with baselines on KITTI and Waymo Dynamic view synthesis in Fig.~\ref{fig:kitti_results} and Fig.~\ref{fig:waymo_results}. 

\begin{figure}[!b]
\centering
    \includegraphics[width=\linewidth]{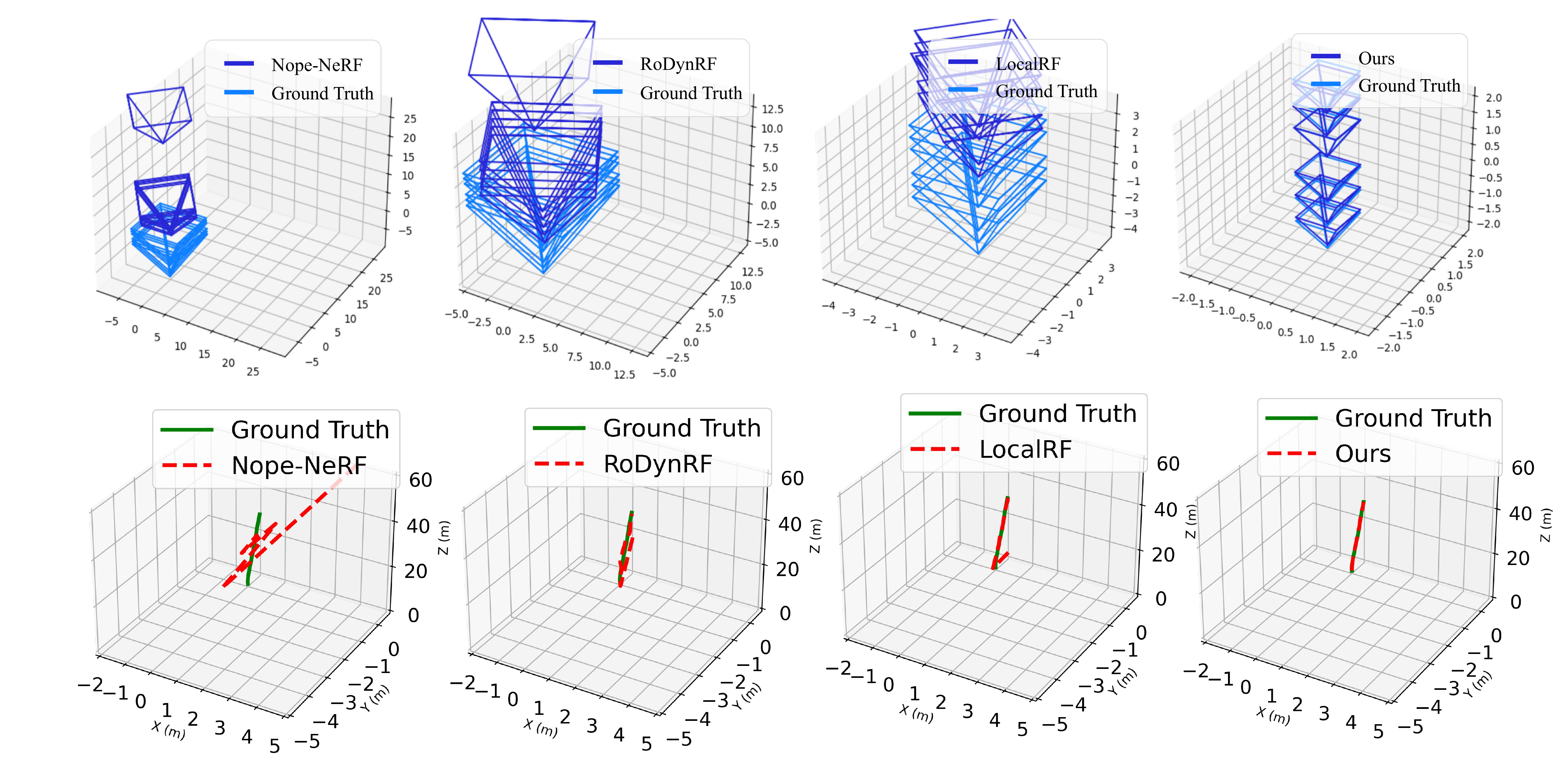}
    \captionof{figure}{Qualitative comparsion for camera pose estimation, including camera position(\textbf{bottom}) and camera orientation(\textbf{top}).}
    \label{fig:pose_visualization}
\end{figure}

\subsection{Evaluation on Camera Poses Estimation}
We follow the post-process of learned camera poses like Nope-NeRF and compare learned poses with ground truth poses of training views. We compare our method with 3 pose-free baselines in two datasets that provide ground truth poses for reference,  including Nope-NeRF\cite{bian2023nope}, RoDynRF\cite{liu2023robust}, and LocalRF\cite{meuleman2023progressively}.  Nope-NeRF and RoDynRF jointly optimize the radiance field and camera pose to achieve satisfactory results in small-scale scenes, especially RoDynRF, trained in 12 images. LocalRF represents a more robust pose estimation using a progressive scheme rather than only a single radiance field. The quantitative camera pose evaluation results are summarized in Tab.~\ref{pose tab1}. Our approach surpasses existing NeRF-based pose-free approaches in inference speed and effectiveness. Estimating accurate camera poses in unbounded urban street scenes relies solely on loss functions, which is challenging. So, we demonstrate the effectiveness and robustness of our method on KITTI and Waymo in Fig.~\ref{fig:pose_visualization}, which present more complex and large camera motions. We also show the quantitative results of estimated camera poses in Tab.~\ref{pose tab1}. We find that the estimated poses of our approach in Waymo are more accurate than others. Even though our estimated camera poses are slightly worse, we still achieve relatively good results due to our robust Gaussian model, depicted in Fig.~\ref{fig:kitti_worse}.

\begin{figure}[h]
\centering
    \includegraphics[width=0.9\linewidth]{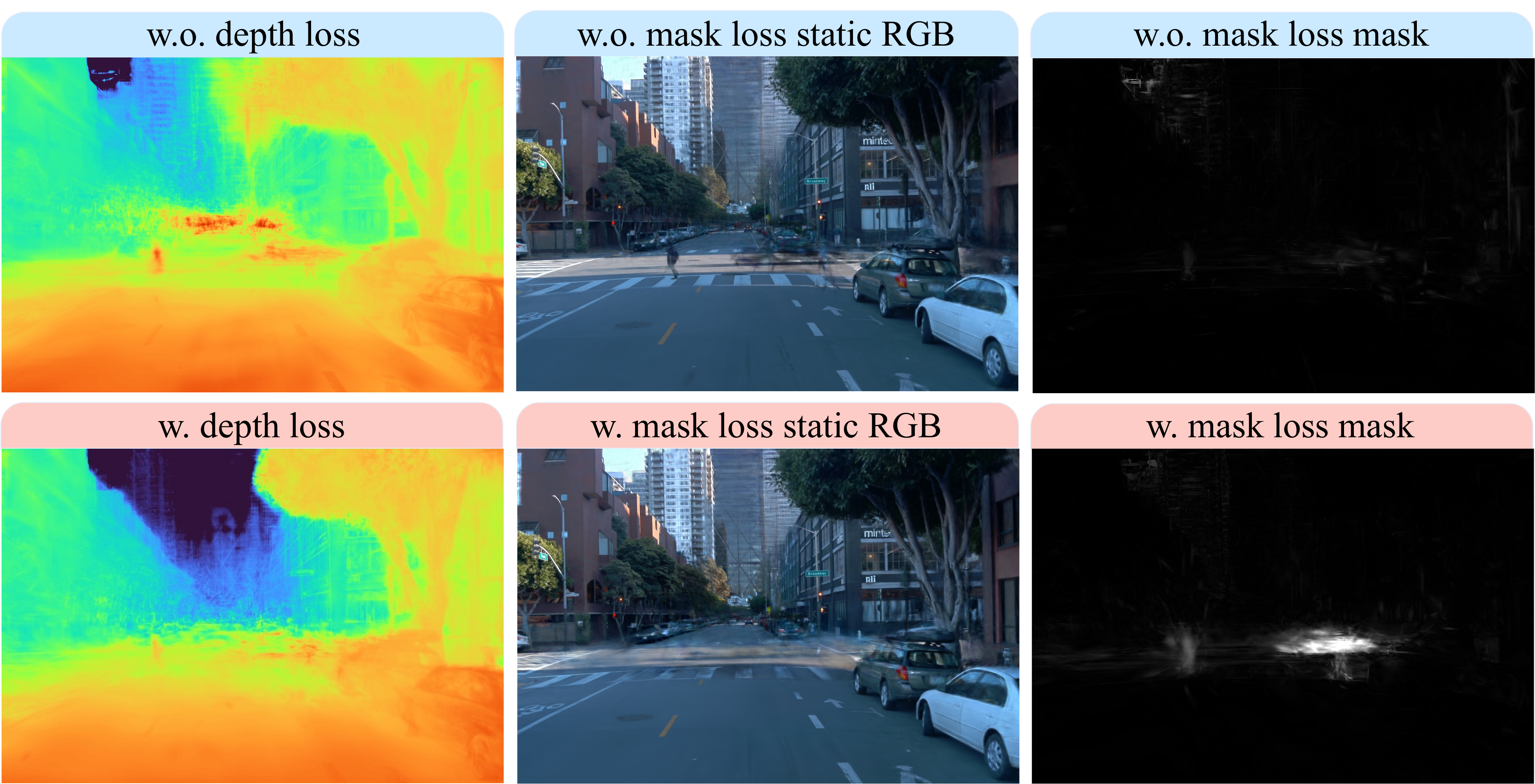}
    \captionof{figure}{Visual ablation results on Waymo Open Dataset. The results indicate that utilizing \(\mathcal{L}_{motion}\) improves the quality of decoupling dynamic and static elements.}
    \label{fig:ablation}
    \vspace{-0.6cm}
\end{figure}
\subsection{Ablation Study}
We conduct an ablation study to indicate the impact of our primary component on the Waymo Open Dataset. Visualization results of the influence of  \(\mathcal{L}_{motion}\) motion mask loss is shown in Fig.~\ref{fig:ablation}. As Tab.~\ref{tab:mask_ablation} shows, we further investigate the impact of the weight of \(\mathcal{L}_{motion}\) on the results. In Fig.~\ref{fig:ablation}, the first column illustrates the effectiveness of using an estimated monocular depth map to supervise our VDG Gaussians set geometric representation; The second column shows the visualization results of dynamic-static decomposition, and the third column demonstrates that we can distinguish dynamic Gaussians points and static Gaussians points.
\begin{table}[H]
    \centering
    \scalebox{0.95}{
        \begin{tabular}{lccc}
            \toprule
            
            & PSNR$\uparrow$  & SSIM$\uparrow$ & LPIPS$\downarrow$ \\
            \midrule
            0.50 * \(\mathcal{L}_{motion}\)& 26.67& 0.848& 0.279\\
            0.05 * \(\mathcal{L}_{motion}\)& 25.79& 0.816& 0.308\\
            0.10 * \(\mathcal{L}_{motion}\)& 26.923& 0.840& 0.278\\
            \bottomrule
        \end{tabular}       
    }
    \caption{Ablation\textbf{ }studies\textbf{ }of  the weight of \textbf{ \(\mathcal{L}_{motion}\) }on the Waymo Open Dataset. }
    \label{tab:mask_ablation}
    \vspace{-3mm}

    \end{table}

\section{Conclusion}
Unlike previous methods that need LiDAR or RGB-D sensors, we propose a novel approach for dynamic driving simulation from multi-camera monocular videos. Our approach utilizes the projection of the estimated monocular depth map rather than random initialization. By integrating self-supervised visual odometry, monocular depth estimation, and \(\mathcal{L}_{motion}\) motion mask loss, our model significantly outperforms the state-of-the-art methods on the Waymo Open Dataset and KITTI benchmark in rendering quality and dynamic-static decomposition. With the proposed model designs, we demonstrate that our model cannot reconstruct precise geometry.

\addtolength{\textheight}{-12cm}   









\bibliographystyle{IEEEtran}
\bibliography{IEEEfull}

\end{document}